\pdfoutput=1
\documentclass[11pt]{article}

\usepackage{acl}

\usepackage{times}
\usepackage{latexsym}

\usepackage[T1]{fontenc}
\usepackage[utf8]{inputenc}

\usepackage{microtype}

\def\ModelName{\textsc{Foam}\xspace}

\usepackage{booktabs} 
\usepackage{multirow}
\usepackage{graphicx}
\usepackage{arydshln}
\usepackage{amsfonts}
\usepackage{amsmath}
\usepackage{amssymb}
\usepackage[ruled,vlined,noend]{algorithm2e}
\DeclareMathOperator*{\argmin}{argmin} 
\usepackage[]{todonotes}

\title{\textsc{Foam}: A Follower-aware Speaker Model For Vision-and-Language Navigation}

\author{Zi-Yi Dou, Nanyun Peng\\
  Department of Computer Science \\ University of California, Los Angeles \\
  {\tt \{zdou,violetpeng\}@cs.ucla.edu }}

\begin{document}
\maketitle
\begin{abstract}
The speaker-follower models have proven to be effective in vision-and-language navigation, where a speaker model is used to synthesize new instructions to augment the training data for a follower navigation model. However, in many of the previous methods, the generated instructions are not directly trained to optimize the performance of the follower. In this paper, we present \textsc{foam}, a \textsc{Fo}llower-\textsc{a}ware speaker \textsc{M}odel that is constantly updated given the follower feedback, so that the generated instructions can be more suitable to the current learning state of the follower. Specifically, we optimize the speaker using a bi-level optimization framework and obtain its training signals by evaluating the follower on labeled data. Experimental results on the Room-to-Room and Room-across-Room datasets demonstrate that our methods can outperform strong baseline models across settings. Analyses also reveal that our generated instructions are of higher quality than the baselines.\footnote{Code is available at \url{https://github.com/PlusLabNLP/follower_aware_speaker}. }
\end{abstract}

\section{Introduction}

The task of vision-and-language navigation (VLN) requires an agent to navigate in a real-world environment given natural language instructions. %\violet{I think the transition to speaker-follower models is a little abrupt. Can we say the major challenge of the problem is the lack of training data(?) and then transit to speaker-follow model to say it helps with the data sparsity issue.}
In VLN, one of the major challenges is the lack of training data. To alleviate the issue, speaker-follower models~\cite{fried2018speaker} have been proposed. Specifically, in the speaker-follower models, an instruction-follower agent is trained to follow a provided natural language instruction to complete a specified goal, and a speaker model learns to model how humans describe routes and synthesize new instructions so as to create more training data for the follower.

\begin{figure}[t]
\centering
\includegraphics[width=0.49\textwidth]{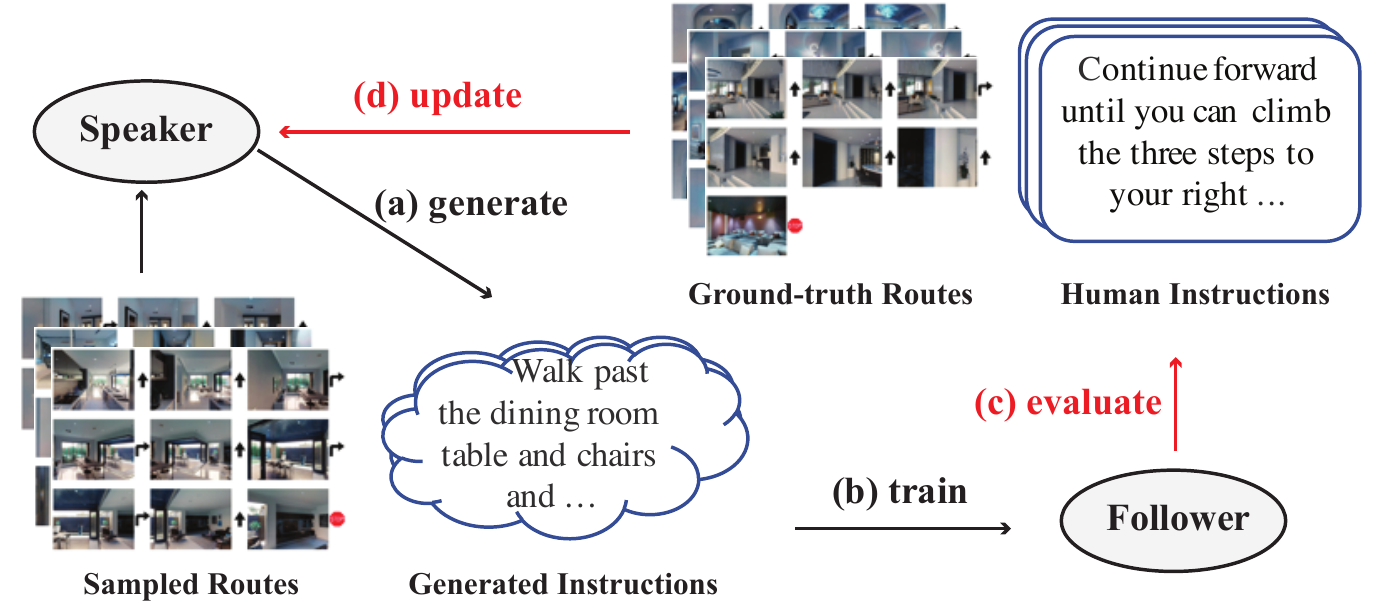}
%%\vspace{-2em}
\caption{Many previous methods use the speaker to generate instructions from sampled routes and train the follower as in the figure from~\citet{fried2018speaker}. \ModelName (in \textcolor{red}{red}) further obtains feedback from the follower on labeled data and updates the speaker accordingly.}
\label{fig:intro}
%%\vspace{-1.5em}
\end{figure} 

While speaker-augmented data is widely used in VLN~\cite{fried2018speaker,wang2019reinforced,ma2019self,tan2019learning,zhu2020vision,hao2020towards,wang2021structured,chen2021history}, most of the previous methods focus on improving the follower navigation model. In contrast, how to improve the speaker model to generate data of higher quality is underexplored. In this line of research,~\citet{fried2018unified} build a pragmatic speaker that can synthesize instructions based on how the follower may interpret the instructions;~\citet{tan2019learning} propose to randomly add noise into the environments when generating instructions, so that the noisy environments can mimic unseen environments and the generated instructions can be more diverse;~\citet{kurita2020generative} propose a generative approach for VLN where a speaker model is trained and the actions of the follower are selected by maximizing the probability of generating the given instruction. %~\citet{liu2021vision} focus on better sampling the trajectories and propose a counterfactual path sampling method. They adversarially sample the most difficult paths for the follower and translate these paths into instructions using the speaker to perform data augmentation. 
%
%However, in most of the existing approaches, 
%In these methods, the speaker model is fixed when training the follower and fails to consider the current learning state of the follower. 

%\zd{create a paragraph here?}
In this paper, we propose a follower-aware speaker model (\ModelName) that optimizes the generated instructions by directly obtaining feedback from the follower so that the generated instructions can be more suitable for the follower. To this end, we frame the idea as a bi-level optimization problem and obtain the feedback signals to improve the speaker based on the follower performance on labeled data. As illustrated in Figure~\ref{fig:intro}, the follower and speaker are trained in an iterative manner: after updating the follower for one step, it is evaluated on a batch of labeled data and the speaker is updated given the performance of the follower. In this way, the speaker is trained to directly optimize the performance of the follower.

Experiments on Room-to-Room~\cite{r2r} and Room-across-Room~\cite{ku2020room} demonstrate strong performance of \ModelName over baselines. Notably, \ModelName can achieve comparable performance to a model pre-trained with over millions of text sentences and image-text pairs. Analyses also reveal that our speaker generates instructions of higher qualities than baselines.

\section{Methods}

We first introduce the background before discussing the details of \ModelName. %discuss our strategies to train the speaker model with signals of follower model's performance. %\violet{I suggest you to create a ``background'' section, where you talk about notations and speaker-follower models. }
\subsection{Background}
\paragraph{Base Settings.} VLN requires an agent to follow a given instruction and find a route in a photo-realistic environment (\textit{e.g.} navigate in indoor living spaces). Formally, in an environment $\mathbf{e}$, the follower $F$ parameterized by $\theta_F$ learns to model the distribution $P(\mathbf{r}|\mathbf{i}; \theta_F)$, where $\mathbf{i}$ and $\mathbf{r}$ denote %are random variables and their corresponding instances are 
instruction and route variables, respectively. %The instruction-follower agent $F$ is typically implemented as a sequence-to-sequence model~\cite{sutskever2014sequence}, where its encoder encodes a sequence of words in the instruction and its decoder outputs route actions sequentially. \violet{can you give an example? maybe put it in the first page? because the output space is quite unclear.}

The training data $\mathcal{D}$ consists of instruction-route pairs from different environments. %, denoted as $D = \{ (i_k, r_k) \}_{k=1}^N$. 
Given a batch of instruction-route pairs $(\mathbf{i}_k, \mathbf{r}_k)$ from $\mathcal{D}$, we train the follower $F$ to minimize the cross-entropy loss between its prediction  $F(\mathbf{i}_k; \theta_F) = P(\hat{\mathbf{r}}|\mathbf{i}_k; \theta_F)$ and the ground-truth label $\mathbf{r}_k$. Here, we denote this supervised loss as $L_l$:  %\violet{looks to me the environment isn't taken into consideration? Also, in general, the concept of ``environment'' is very vague -- i don't understand what does it mean and how it affect your learning.}\zd{i've removed the term `environment`}
%\violet{this is confusing as you constantly switching from $R$, $R_i$, and seem to use them inter-changeably (similarly for $I$). I suggest you to use upper case to denote a set and lower case to denote it's number. And you can use subscription for the members. E.g., $R$ and $r_i$}
%\zd{maybe change $\mathbf{r}$ to $\mathbf{r}_k$ etc. in the equations?}

{\small 
\begin{equation}
\label{eqn:forward}
\min_{\theta_F} L_l(\theta_F) = \mathbb{E}_{(\mathbf{i}_k, \mathbf{r}_k) \sim \mathcal{D}} [\textrm{CE}(\mathbf{r}_k, F(\mathbf{i}_k; \theta_F)))].
\end{equation}
}

\paragraph{Speaker-Follower Models.} \citet{fried2018speaker} propose to train a speaker model $S$ parameterized by $\theta_S$ that models the distribution of $P(\mathbf{i}|\mathbf{r}; \theta_S)$. As in Figure~\ref{fig:intro}, with the speaker, we can perform back translation~\cite{sennrich2016improving} on randomly sampled routes $\hat{\mathbf{r}}$ from the training environments $\mathcal{E}$ %\violet{zi-yi, can you add a footnote to explain the environment and a sampled route? Just as you explained in the slack. \zd{explain environment in the first paragraph and also refer to Figure 1 for sampled paths}} 
for data augmentation. Specifically, we first train the speaker $S$ on the same training data as the follower. Then, given a batch of sampled route $\hat{\mathbf{r}}_k \sim \mathcal{E}$, we synthesize their human-like textual instructions $\hat{\mathbf{i}}_k=S(\hat{\mathbf{r}}_k; \theta_S)$. Afterwards, the synthesized training instances $(\hat{\mathbf{i}}_k, \hat{\mathbf{r}}_k)$ are used to update $F$. Here, we denote this loss as $L_u$:

{\small \begin{equation}
\label{eqn:back}
\begin{aligned}
      \min_{\theta_F}  L_u(\theta_F, \theta_S) & = \mathbb{E}_{(\hat{\mathbf{i}}_k, \hat{\mathbf{r}}_k) \sim \mathcal{E}} [\textrm{CE}(\hat{\mathbf{r}}_k, F(\hat{\mathbf{i}}_k; \theta_F))] \\
      & = \mathbb{E}_{\hat{\mathbf{r}}_k \sim \mathcal{E}} [\textrm{CE}(\hat{\mathbf{r}}_k, F(S(\hat{\mathbf{r}}_k; \theta_S); \theta_F))].
\end{aligned}
\end{equation}
}

\subsection{Optimizing the Speaker}
\label{sec:speaker}
As we can see from Equation~\ref{eqn:back}, the resulting follower parameters $\theta_F^*$ depends on the speaker parameters $\theta_S$, and we can express the dependency as $\theta_F^*(\theta_S)$. However, existing speaker-follower models fail to incorporate $\theta_S$ into the optimization process and $\theta_S$ is always fixed during training.

\paragraph{Formulation.} In this paper, we propose to optimize the parameters of both the follower and speaker during back translation.
Specifically, taking inspirations from~\citet{pham2021meta,pham2020meta}, we optimize the speaker based on the performance of the follower on the labeled training data, which can be expressed as:

{\small
\begin{equation}
\label{eqn:meta}
\begin{aligned}
\min_{\theta_S}    \,\, &
L_l(\theta_F^*(\theta_S)), \\
\textrm{where }  \theta_F^*(\theta_S) &= \argmin_{\theta_F} L_u(\theta_F, \theta_S).
\end{aligned}
\end{equation}
}

The motivation of Equation~\ref{eqn:meta} is that while the speaker-augmented data can provide additional supervisions for the follower, the main objective of the speaker is to make the follower better follow human instructions, thus we should focus on minimizing follower's loss on the labeled training data.

\paragraph{Approximation.} Following previous work in bi-level optimization~\cite{finn2017model,liu2018darts,pham2021meta,pham2020meta}, we can approximate $\argmin$ with one-step gradient update and alternatively update the parameters $\theta_F$ and $\theta_S$.

Specifically, at training step $t$, we first sample a batch of routes and synthesize their instructions using the speaker $S$. The generated data is used to update the follower:

{\small
\begin{equation}
    \theta_F^{t} = \theta_{F}^{t-1} - \eta_F \nabla_{\theta_F} L_u(\theta_F^{t-1}, \theta_S^{t-1}),
\end{equation}}
where $\eta_F$ is the learning rate.

Then, the speaker is updated to optimize the objective $L_l(\theta_{F}^{t}))$ with

{\small \begin{equation}
\label{eqn:speaker-update}
    \theta_S^{t}  = \theta_S^{t-1} - \eta_S  \nabla_{\theta_S} L_l(\theta_{F}^{t}(\theta_S)).
\end{equation}}

We can approximate the gradient $\nabla_{\theta_S} L_l(\theta_{F}^{t}))$ (derivation details in Appendix~\ref{sec:app-speaker}) with 

{\tiny \begin{equation} 
\label{eqn:speaker-gradient}
-[\nabla_{\theta_F} L_l(\theta_F^t)^{\mathrm{T}} \nabla_{\theta_F} L_u(\theta_F^{t-1}, \theta_S^{t-1})  ] \nabla_{\theta_S} \log P(\hat{\mathbf{i}}_k|\hat{\mathbf{r}}_k; \theta_S^{t-1}).
\end{equation}}
We can see that this equation resembles REINFORCE~\cite{williams1992simple} in reinforcement learning. Therefore, this algorithm can also be interpreted as treating the similarity in the gradients of the follower model on the labeled data and on the augmented data as rewards, and update the speaker model using reinforcement learning.

\begin{table*}[t]
  \centering
  \small
  \begin{tabular}{@{}lccccccccccc@{}}
    \toprule
     \multirow{2}[4]{*}{\textbf{Model}} & \multicolumn{3}{c}{\textbf{Val-Seen}} & \multicolumn{3}{c}{\textbf{Val-Unseen}} & \multicolumn{3}{c}{\textbf{Test}}  \\
\cmidrule(lr){2-4}\cmidrule(l){5-7}  \cmidrule(l){8-10}         &   \textbf{SR$\uparrow$} & \textbf{SPL$\uparrow$} & \textbf{NE$\downarrow$} &    \textbf{SR$\uparrow$} & \textbf{SPL$\uparrow$} & \textbf{NE$\downarrow$} &    \textbf{SR$\uparrow$} & \textbf{SPL$\uparrow$} & \textbf{NE$\downarrow$}   \\
    \midrule
      \multicolumn{4}{l}{\textit{{Previous Work }} } \\
    \midrule
    %RCM+SIL (train) \cite{wang2019reinforced} & 66.7 & - & 3.53 & 42.8 &- &6.09 & 43.0 & 38 & 6.12 \\
    EnvDrop-ResNet \cite{tan2019learning} &  62.1 & 59 & 3.99 & 52.2 & 48 & 5.22 & 51.5 & 47 & - \\
    AuxRN~\cite{zhu2020vision} & 70 & \bf 67 & 3.33 & 55 & 50 & 4.71 & 55 & 51 & 5.15\\
    RelGraph~\cite{hong2020language} & 67 & 65 & 3.47 & 57 & 53 & 4.73 & 55 & 52 & 4.75 \\
    EnvDrop-CLIP-ResNet \cite{shen2021much} &  - & - & - & - & - & - & 59.2 & 53 & -\\
    \midrule
      \multicolumn{4}{l}{\textit{{Our Implementations}} } \\
      \midrule
      Base Follower-CLIP-ViT  & 60.5 & 56.6 & 3.97 & 54.9 & 49.3  & 4.81 & - & - & -  \\
      EnvDrop-CLIP-ViT &  66.1 & 61.7 & 3.61 & 59.2 & 52.4 & 4.31 & 60.0 & 53.9 & 4.38 \\
      \ModelName-CLIP-ViT  & \bf 70.8 & \bf 66.6 & \bf 3.25 & \bf 61.6 & \bf 55.1 & \bf 4.18 & \bf 62.2 & \bf 56.2 & \bf 4.09\\
    \bottomrule
    \end{tabular}%
  \caption{Results on Room-to-Room. We report success rates (SR), success rates weighted by path length (SPL), navigation error (NE). The best scores are in \textbf{bold}. We implement the models based on CLIP-ViT which is stronger than ResNets (row 6 vs. row 1/4). `Base Follower' is our follower model pre-trained without using the speaker-augmented data. `EnvDrop' is the best existing speaker-follower baseline.}%  \textcolor{red}{can you add the $\uparrow$ and $\downarrow$ to the table?}\zd{done}}
  \label{tab:r2r}
  %\vspace{-.5em}
\end{table*}

\paragraph{End-to-End Reconstruction Loss.} \label{sec:ablation} In this paper, we also propose to add a reconstruction loss for the speaker. Concretely, we compute the gradient of Equation~\ref{eqn:back} with respect to the speaker parameter $\theta_{S}$ using straight-through estimator, denoted as $\nabla_{\theta_S} L_u(\theta_F, \theta_S)$, and then update the speaker in an end-to-end manner. 

To sum up, in \ModelName, the final gradient of the speaker is computed based on both the reconstruction loss (Equation~\ref{eqn:back}) and the bi-level optimization loss (Equation~\ref{eqn:speaker-gradient}), and we will perform ablations on the two objectives in the experiment section.

\begin{table*}[t]
  \centering
  \small
  \renewcommand{\tabcolsep}{0.2mm}
  \begin{tabular}{@{}l@{\ \ }cccc@{\ \ }cccc@{\ \ }cccc@{\ \ }cccc@{}} %ccccccccccc
    \toprule
     \multirow{2}[4]{*}{\textbf{Model}} & \multicolumn{4}{c}{\textbf{Val-Unseen-English}} & \multicolumn{4}{c}{\textbf{Val-Unseen-Hindi}} & \multicolumn{4}{c}{\textbf{Val-Unseen-Telugu}}  & \multicolumn{4}{c@{}}{\textbf{Test}} \\
\cmidrule(lr){2-5}\cmidrule(l){6-9}  \cmidrule(l){10-13}  \cmidrule(l){14-17}        &   \textbf{SR$\uparrow$} & \textbf{SPL$\uparrow$}  &  \textbf{sDTW$\uparrow$} &  \textbf{nDTW$\uparrow$}  &  \textbf{SR$\uparrow$} & \textbf{SPL$\uparrow$}  &  \textbf{sDTW$\uparrow$} &  \textbf{nDTW$\uparrow$} &    \textbf{SR$\uparrow$} & \textbf{SPL$\uparrow$}   &  \textbf{sDTW$\uparrow$} &  \textbf{nDTW$\uparrow$}   &    \textbf{SR$\uparrow$} & \textbf{SPL$\uparrow$}   &  \textbf{sDTW$\uparrow$} &  \textbf{nDTW$\uparrow$}  \\
    \midrule
      Base  &  40.7 & 36.4 & 33.5 & 52.8 & \bf 46.8 & 41.5 &  38.5 & 56.1 & 42.6 & 38.3 & 35.1 & 54.6 & 39.1 & 35.2 & 32.7 & \bf 49.7 \\
      EnvDrop & 42.4 & 38.3 & 35.5    & 53.9 & 46.5 & 41.5 & 38.5 & 56.0 & 44.4 & 39.3 & 36.5 & \bf 54.8 & \bf 41.2 & \bf 36.3 & \bf 33.6 & 48.8  \\
      \ModelName & \bf 42.8 &  \bf38.7 & \bf 35.6 & \bf 54.1 & 46.7 & \bf 41.8 & \bf 38.6 & \bf 56.5 & \bf 45.6 & \bf 39.7 & \bf 37.0 & 54.4 & \bf 41.2 & 36.2 &\bf  33.6 & 49.3 \\
    \bottomrule
    \end{tabular}%
  \caption{Results on Room-across-Room. We report success rates (SR), success rates weighted by path length (SPL), success rates weighted by dynamic time warping (sDTW), normalized dynamic
time warping (nDTW). The best scores are in \textbf{bold}. }
  \label{tab:rxr}
  %\vspace{-1em}
\end{table*}

\begin{table*}[t]
  \centering
  \small
  \begin{tabular}{@{}cccccccccccc@{}}
    \toprule
     \multirow{2}[4]{*}{\textbf{Follower }} & \multirow{2}[4]{*}{\textbf{Speaker  }} & \multicolumn{3}{c}{\textbf{ Pre-exploration}} &  \multicolumn{3}{c}{\textbf{Beam Search} } & \\
     \cmidrule(lr){3-5}\cmidrule(l){6-8}  
     && \multicolumn{1}{c}{\textbf{Val-Seen}} & \multicolumn{1}{c}{\textbf{Val-Unseen}} & \multicolumn{1}{c}{\textbf{Test}}   & \multicolumn{1}{c}{\textbf{Val-Seen}} & \multicolumn{1}{c}{\textbf{Val-Unseen}} & \multicolumn{1}{c}{\textbf{Test}}\\
    \midrule
    EnvDrop & EnvDrop  & 66.9 & 64.2 & - & 74.9 & 68.4 & -\\
    \ModelName & EnvDrop  & 70.2 & 66.0 & - & 77.0 & 70.6 & -\\
    \ModelName & \ModelName & 70.6 & 66.5 & 68.4 & 78.1 & 72.1 & 72.2 \\
    \bottomrule
    \end{tabular}%
  \caption{Success rates of different configurations of the speaker-follower models in pre-exploration and beam search settings on Room-to-Room. The best configuration is using both \textit{our} follower and \textit{our} speaker models. }
  \label{tab:beam-pre}
  %\vspace{-1em}
\end{table*}

\section{Experiments}
%We then perform experiments in this section.
%\subsection{Experimental Setup}
\paragraph{Datasets.} We evaluate the models on the Room-to-Room (R2R)~\cite{r2r} and Room-across-Room (RxR)~\cite{ku2020room} datasets. The R2R dataset consists of 7,189 paths, and each path has 3 English instructions with an average length of 29. R2R is split into training, validation, and test sets. The validation set is split into \textit{val-seen}, where paths are sampled from environments seen during training, and \textit{val-unseen}, where paths are sampled from environments that are not seen during training. The paths in the test set are from new environments unseen in the training and validation sets. The RxR dataset follows the same environment division as R2R and there are 16,522 paths in total. %\violet{may want to give more details about the extension: how? by translation to other languages, or extend and then translate? } \zd{changed the sentence here} 
The instructions have an average length of 78 and are in three languages, including English, Hindi, and Telugu. %The RxR dataset .

\paragraph{Evaluation Metrics.} Our primary metric is success rate (SR), and we also report navigation error (NE), success rate weighted by path length (SPL) on R2R. Following the suggestion in~\citet{ku2020room}, we also report normalized dynamic
time warping (nDTW) and success rate weighted by dynamic time warping (sDTW)~\cite{Magalhes2019EffectiveAG} on RxR.

\paragraph{Implementation Details.} Following EnvDrop~\cite{tan2019learning},we build our speaker and follower based on LSTM~\cite{hochreiter1997long} and environmental dropout is used during back-translation.
The follower is pre-trained with imitation and reinforcement learning, and the speaker is pre-trained with maximum likelihood training. Here, we refer to this pre-trained follower as \textbf{base follower}. The two models are pre-trained for 80k steps on R2R and 200k steps on RxR, and then trained with our method until the 300k-th iteration. We perform environmental dropout during training as in~\citet{tan2019learning}, and also use their 176,776 paths randomly sampled from seen environments for back translation. Different from~\citet{tan2019learning},  we use CLIP-ViT-224/16~\cite{radford2021learning} to extract vision features as CLIP vision encoders can be beneficial for VLN models~\cite{shen2021much} and we demonstrate that using CLIP vision encoder can obtain better performance than ResNet-based models in the following parts. We compute the cosine similarities between gradients for Equation~\ref{eqn:speaker-gradient} following~\citet{pham2020meta,pham2021meta} and also perform the same weighting for the reconstruction loss. Each training takes about 3 days on 1 NVIDIA V100 GPU to finish. We report numbers of a single run for evaluations.

\subsection{Main Results}
\paragraph{Room-to-Room.} We report the main results on R2R in Table~\ref{tab:r2r}. We can see that our implementation of the baseline EnvDrop model is better than the previous work because of the stronger vision encoder we use. Based on the strong baseline, our model achieves further improvements on both validation and test sets, outperforming EnvDrop by 2.2\% in the success rate on the R2R test dataset, suggesting that our framework is indeed effective.

\paragraph{Room-across-Room.} We report the main results on R2R in Table~\ref{tab:rxr}. From the table, we can see that the improvements of our framework are not as good on the RxR dataset, possibly because the instructions are much longer and thus it is hard to train a good speaker. Specifically, we find that the baseline speaker can only achieve a BLEU score of 7.4 on the English validation set on RxR (compared with over 30 BLEU scores on R2R as in Appendix~\ref{app:bleu}), which leads to noisy augmented data and can impact the performance of speaker-follower models.

\begin{figure}[t]
\centering
\includegraphics[width=0.49\textwidth]{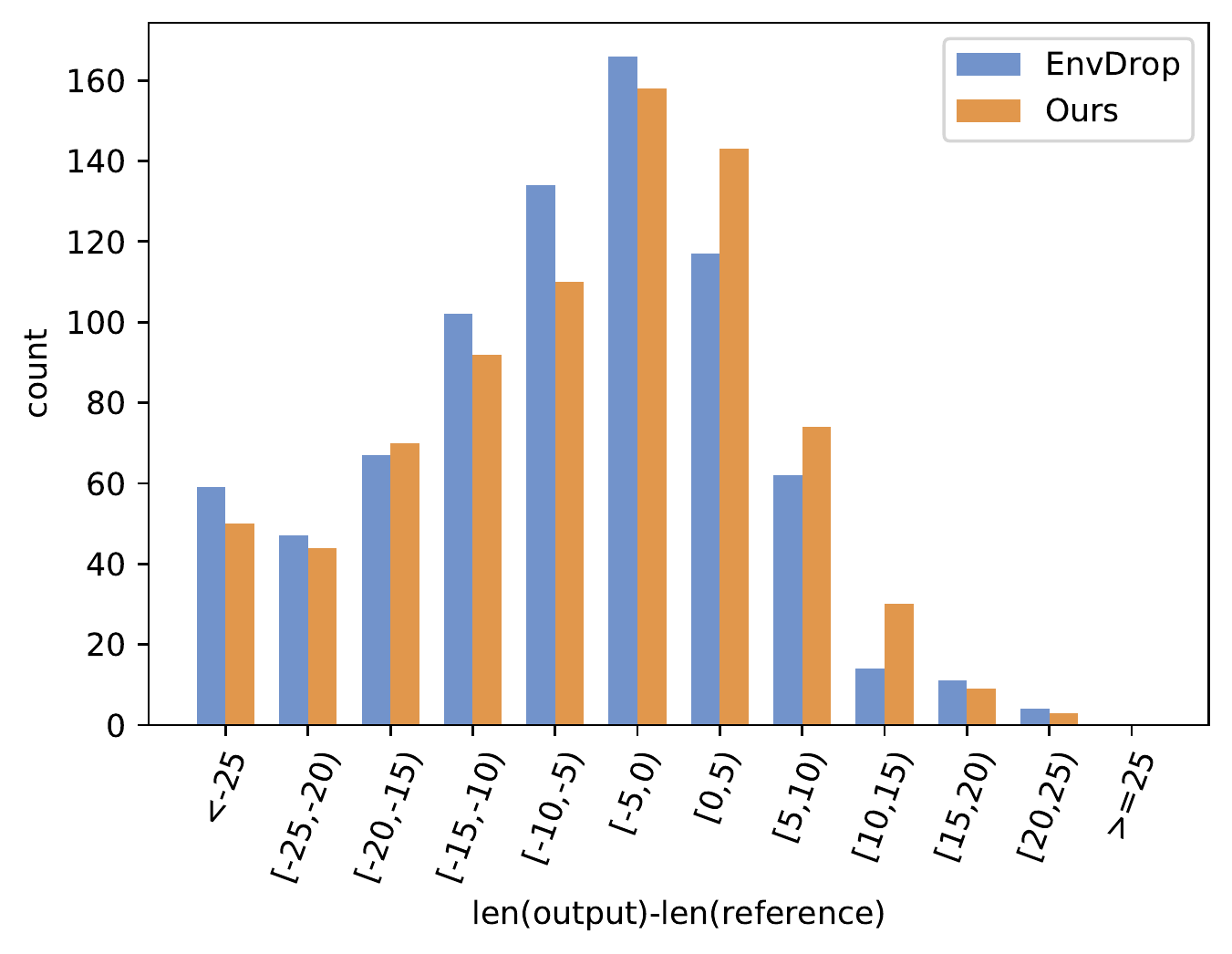}
%\vspace{-1em}
\caption{A histogram of the differences in length between the reference and model outputs. Baseline can often generate shorter instructions than the references, but our method can alleviate the issue.}
\label{fig:len}
%\vspace{-0.5em}
\end{figure}

\subsection{Analysis}
We then perform analyses to gain more insights regarding our models:

\paragraph{Pre-exploration and Beam Search.} We perform experiments in both pre-exploration and beam search settings following previous work~\cite{tan2019learning}. Because both the speaker and follower are used in the two settings, the evaluation results  can reflect the quality of both of the models. As shown in Table~\ref{tab:beam-pre}, we find that the best configuration is using \textit{our} follower and \textit{our} speaker, suggesting that both our follower and speaker are more suitable for VLN than the baselines. Notably, in the beam search setting, our model can achieve a success rate of 72.2\%, which is comparable to VLN-BERT~\cite{majumdar2020improving} that achieves a success rate of 73\% and is pre-trained with over millions of text sentences and image-text pairs.

\begin{table}[t]
  \centering
  \small
  \begin{tabular}{@{}lccccccccccc@{}}
    \toprule
       \multirow{2}[4]{*}{\textbf{Model}} & \multicolumn{3}{c}{\textbf{Val-Seen}} & \multicolumn{3}{c}{\textbf{Val-Unseen}} \\
\cmidrule(lr){2-4}\cmidrule(l){5-7}  \cmidrule(l){8-10}         &   \textbf{SR} & \textbf{SPL} & \textbf{NE} &    \textbf{SR} & \textbf{SPL} & \textbf{NE} \\
\midrule
    \ModelName & 70.8 & 66.6 & 3.25 & 61.6 & 55.1 & 4.18  \\
    -Recon. & 68.9 & 63.5 & 3.33 &  60.2 & 53.1 & 4.30\\
    -Bi-level &  69.6 & 65.3 &  3.33 &  60.7 & 54.6 & 4.27 \\
    \bottomrule
    \end{tabular}%
  \caption{Ablation studies on our proposed objectives. Our reconstruction loss and bi-level optimization loss are complementary to each other and ablating either one of them can lead to degraded performance.}
  \label{tab:ablation}
  %\vspace{-0.5em}
\end{table}

\paragraph{Generated Instructions.} 
The previous pre-exploration and beam search results well indicate that our generated instructions are more suitable for our follower, suggesting the effectiveness of our framework. In this paragraph, we also compare the generated instructions with the reference instructions. In Figure~\ref{fig:len}, we plot the histogram of length differences between the reference
sentences and the generated instructions using \textit{compare-mt}~\cite{neubig2019compare}. The figure suggests that the baseline model can often generate shorter instructions than the references, but our method can alleviate this issue, indicating that our methods can indeed improve the speaker quality during training. We also find that our generated instructions are quantitatively and qualitatively better than the baseline using automatic evaluations as in Appendix~\ref{app:bleu}.

\paragraph{Ablation Studies.} As mentioned in Section~\ref{sec:ablation}, we perform ablation studies on both of our proposed objectives, namely the bi-level optimization loss (Equation~\ref{eqn:speaker-update}) and reconstruction loss. As shown in Table~\ref{tab:ablation}, ablating either of the objectives can lead to degraded performance on the R2R validation sets, indicating that both the objectives can improve the model performance and they are complementary to each other.

\section{Related Work}
We overview two lines of related work:
\paragraph{Vision-and-Language Navigation.} Training embodied navigation agents has been an increasingly active research area~\cite{anderson2018evaluation,r2r,chen2019touchdown,ku2020room,shridhar2020alfred,padmakumar2021teach}.~\citet{fried2018speaker} propose to augment the training data with the speaker-follower models, which is improve by~\citet{tan2019learning} who add noise into the environments so that the speaker can generate more diverse instructions.~\citet{zhao2021evaluation} propose methods to measure the quality of the generated instructions and filter noisy samples.~\citet{liu2021vision} propose to adversarially sample the most difficult paths for the follower and translate these paths into instructions using the speaker for data augmentation.  While using the speaker-augmented data has been widely used in VLN, most of the existing work has been focused on improving the follower navigation model~\cite{wang2018look,li2019robust,zhu2020babywalk}. For example, the self-monitoring agent~\cite{ma2019self} improves cross-modal alignment through a visual-text co-grounding module and a progress monitor;~\citet{zhu2020vision} propose to utilize four self-supervised auxiliary tasks that can provide additional training signals for the agent. Most similar to our work,~\citet{fried2018unified} build a speaker that reason about how the instructions may be interpreted;~\citet{kurita2020generative} propose a generative approach where a speaker model is trained to model the probability of an instructions given actions, and the follower chooses actions that maximize this probability.

\paragraph{Bi-level Optimization.} Bi-level optimization algorithms have been widely applied in various fields, such as learning initialization parameters~\cite{finn2017model}, neural architecture search~\cite{liu2018darts}, re-weighting training data~\cite{wang2020optimizing}. Our method takes inspirations from~\cite{pham2021meta}, which is applied in pseudo labeling and optimizes the teacher parameters given the student feedback. Similar techniques have also been used in machine translation~\cite{pham2020meta}, where a meta-validation set is constructed to evaluate the model performance and provide feedback. %In this paper, we adopt the techniques in vision-and-language navigation and does not construct a meta-validation set for simplicity.

\section{Conclusions}
In this paper, we propose the \ModelName model where we improve the speaker-follower model in vision-and-language navigation by constantly updating the speaker given the follower feedback during training. We frame the idea as a bi-level optimization problem and obtain the feedback signal based on the performance of the follower on labeled data. Experimental results on Room-to-Room and Room-across-Room datasets demonstrate that our method can outperform strong VLN baselines in different settings. Analyses also suggest that the quality of our speaker model is indeed improved during training. Future directions include testing our method on more datasets and investigating more options on the feedback signals.

\section*{Acknowledgement}
We would like to thank the anonymous reviewers for valuable suggestions and Te-Lin Wu for helpful discussions. 
This work is supported in part by the DARPA Machine Common Sense (MCS) program under Cooperative Agreement N66001-19-2-4032 and NIH R01HL152270. 
%The views and the conclusions
% of this paper are those of the authors and do not
% reflect the official policy or position of DARPA.

\bibliography{custom}
\bibliographystyle{acl_natbib}

\appendix

\section{Derivation of the Speaker Gradient}
\label{sec:app-speaker}

As shown in Section~\ref{sec:speaker}, at training step $t$, we update the follower according to:
\begin{equation}
    \theta_F^{t} = \theta_{F}^{t-1} - \eta_F \nabla_{\theta_F} L_u(\theta_F^{t-1}, \theta_S^{t-1}).
\end{equation}
We then derive the speaker gradient following previous work~\cite{pham2020meta,pham2021meta}. We define the \textit{expected} parameters of the follower as $\bar{\theta}_F^{t}$:
\begin{equation}
\small
    \bar{\theta}_F^{t} = \mathbb{E}_{\hat{\mathbf{r}}_k \sim \mathcal{E}, \hat{\mathbf{i}}_k\sim P(\mathbf{i}|\hat{\mathbf{r}}_k; \theta_S^{t-1})} [ \theta_{F}^{t-1} - \eta_F \nabla_{\theta_F} L_u(\theta_F^{t-1}, \theta_S^{t-1})].
\end{equation}

Then, using the chain rule, we can obtain
\begin{equation}
\label{eqn:chain}
    \nabla_{\theta_S} L_l= \frac{\partial L_l}{\partial \bar{\theta}_F^{t}  } \frac{\partial\bar{\theta}_F^{t} }{\partial \theta_S},
\end{equation}
where the first term can be approximated with $\frac{\partial L_l}{\partial {\theta}_F^{t}  }$.
Then, for the second term, we have
\begin{equation}
\small
    \frac{\partial\bar{\theta}_F^{t} }{\partial \theta_S} = \frac{\partial }{\partial \theta_S} \mathbb{E}_{\hat{\mathbf{r}}_k\sim \mathcal{E}, \hat{\mathbf{i}}_k\sim P(\mathbf{i}|\hat{\mathbf{r}}_k)} [ \theta_{F}^{t-1} - \eta_F \nabla_{\theta_F} L_u(\theta_F^{t-1}, \theta_S^{t-1})].
\end{equation}

We can assume that $\theta_{F}^{t-1}$ does not depend on $\theta_S$ with Markov assumption~\cite{pham2021meta}, and apply the REINFORCE~\cite{williams1992simple} equation on the second term:
\begin{equation}
\tiny
\label{eqn:final}
\begin{aligned}
        & \frac{\partial\bar{\theta}_F^{t} }{\partial \theta_S}  = \frac{\partial }{\partial \theta_S} \mathbb{E}_{\hat{\mathbf{r}}_k\sim \mathcal{E}, \hat{\mathbf{i}}_k\sim P(\mathbf{i}|\hat{\mathbf{r}}_k)} [- \eta_F \nabla_{\theta_F} L_u(\theta_F^{t-1}, \theta_S^{t-1})] \\
        & = -\eta_F  \mathbb{E}_{\hat{\mathbf{r}}_k\sim \mathcal{E}, \hat{\mathbf{i}}_k\sim P(\mathbf{i}|\hat{\mathbf{r}}_k)} [\nabla_{\theta_F} L_u(\theta_F^{t-1}, \theta_S^{t-1}) \frac{\partial}{\partial \theta_S} \log P(\hat{\mathbf{i}}_k|\hat{\mathbf{r}}_k; \theta_S^{t-1})], \\
\end{aligned}
\end{equation}

Using Monte Carlo approximation to approximate terms in Equation~\ref{eqn:final} using a batch of samples and substituting the result into Equation~\ref{eqn:chain}, we can get 
\begin{equation}
\tiny
    \nabla_{\theta_S} L_l= -\eta_F[\nabla_{\theta_F} L_l(\theta_F^t)^{\mathrm{T}} \nabla_{\theta_F} L_u(\theta_F^{t-1}, \theta_S^{t-1})  ] \nabla_{\theta_S} \log P(\hat{\mathbf{i}}_k|\hat{\mathbf{r}}_k; \theta_S^{t-1}).
\end{equation}

Note that here $\eta_F$ is a hyper-parameter and can be incorporated into the learning rate of the speaker $\eta_S$, thus we remove this term in Section~\ref{sec:speaker} and our derivation is complete.

\section{Evaluations of the Generated Instructions}
\label{app:bleu}
\begin{table}[t]
  \centering
  \small
  \begin{tabular}{@{}lccccccccccc@{}}
    \toprule
     \multicolumn{1}{l}{\textbf{Model}} & \multicolumn{1}{c}{\textbf{Train}} & \multicolumn{1}{c}{\textbf{Val-Seen}} & \multicolumn{1}{c}{\textbf{Val-Unseen}} \\
    \midrule
      \multicolumn{4}{l}{\textit{{BLEU}} } \\
    \midrule
    EnvDrop & 38.16 & 32.42 & 31.13 \\
    \ModelName  & 39.66 & 33.11 & 31.10\\
    \midrule
      \multicolumn{4}{l}{\textit{{BERTScore}} } \\
    \midrule
    EnvDrop & 91.64 & 91.08 & 91.04 \\
    \ModelName  &  91.79 & 91.08 & 91.10 \\
    \bottomrule
    \end{tabular}%
  \caption{Automatic evaluations of the generated instructions. The instructions generated by our model can obtain higher BLEU and BERTScore than the baseline.}
  \label{tab:bleu}
\end{table}

\paragraph{Automatic Evaluations.} As in Table~\ref{tab:bleu}, we measure the quality of the generated instructions in BLEU~\cite{papineni2002bleu} and BERTScore~\cite{zhang2019bertscore}. We find that our speaker can generate instructions of higher qualities according to the two metrics.

  \begin{table}[th]
  \centering
  \small
  \begin{tabular}{@{}l@{\ \ }p{6cm}@{}}
  \toprule
  \bf Method & \bf Instruction \\
  \midrule
  Reference & walk downstairs and outside . stop in the outhouse through the door on the right .\\
  EnvDrop  & go down the stairs and turn right . go down the hallway and stop in front of the door .\\
  \ModelName & go down the stairs and turn right . go down the hallway and go through the door on the right .\\
  \midrule
   Reference & turn left and take a right at the table . take a left at the painting and then take your first right . wait next to the exercise equipment . \\
   EnvDrop & walk past the dining room table and chairs and turn left . walk past the table and chairs and turn right . walk into the room and stop .\\
   \ModelName & walk past the dining room table and chairs and turn left . walk past the table and chairs and turn right . walk into the room and turn right . stop in front of the exercise bike .\\
 \bottomrule
    \end{tabular}
    \caption{ \label{tab:examples} Examples of the generated instructions. Our generated instructions are generally longer and more accurate compared with the baseline.}
  \end{table}
  
\paragraph{Qualitative Examples.} As in Table~\ref{tab:examples}, we also find that after training the speaker using our method, the generated instructions are generally longer than the baseline and are more accurate compared with the references.

\section{License}
We evaluate our models on the Room-to-Room (R2R)~\cite{r2r} and Room-across-Room (RxR)~\cite{ku2020room} datasets based on Matterport3D~\cite{Matterport3D}. The datasets are released under the Matterport3D Terms of Use.\footnote{\url{http://dovahkiin.stanford.edu/matterport/public/MP_TOS.pdf}} The datasets do not contain any information that names or uniquely identifies individual people or offensive content. Our code is based on EnvDrop that is released under the MIT license.\footnote{\url{https://github.com/airsplay/R2R-EnvDrop/blob/master/LICENSE}} We use the datasets and code for research purposes, which is consistent with their intended use. 

\section{Limitations and Potential Risks}
As in the experiments, our models may not work well when the instructions are long and it is hard to train a reasonable speaker model. Also, our model requires fine-tuning the speaker during training based on the feedback of the follower, which introduces additional training costs to the model. In addition, the datasets we use in the paper may make our model biased towards environments of American buildings.

\end{document}